\documentclass[conference]{IEEEtran}
\IEEEoverridecommandlockouts   

\usepackage{cite}
\usepackage{amsmath,amssymb,amsfonts}

\usepackage{algorithm}
\usepackage{algorithmic}
\usepackage{graphicx}
\usepackage{textcomp}
\usepackage{xcolor}
\usepackage{multirow}
\usepackage[normalem]{ulem} 

\begin{document}

\title{Real-Time Compliance and Position Control of a Hyper-redundant Soft Robotic Arm}

\author{%
    Runze~Zuo$^{*}$,
    Tianhua~Zou,
    Naike~Wu,
    Mingyuan~Li,
    and~Daniel~Bruder%
    \thanks{All authors are with the Department of Mechanical Engineering,
        University of Michigan, Ann Arbor, MI 48109, USA.}%
    \thanks{$^{*}$R.~Zuo is the corresponding author
        (e-mail: \texttt{zuorunze@umich.edu}).}%
}

\maketitle

\begin{abstract}
Robots working in unstructured or partially unobservable environments must combine accurate motion
with physical compliance that can passively correct contact misalignment. Soft robots provide this
compliance but have struggled to precisely control their tip compliance and position. This paper presents a robot architecture designed around that control problem: a 7-link arm whose six articulated joints provide twelve independently driven revolute axes,
each actuated by an antagonistic pair of pneumatic muscles, so that every axis
can simultaneously change its angle and linearly adjust its stiffness.
The rigid articulated backbone makes the tip compliance and position of the arm
predictable enough to be commanded quantitatively in real time.
The robot employs a unified iterative inverse-kinematics and
inverse-compliance controller to achieve simultaneous, quantitative control of both compliance and
position. The task-space compliance and kinematics models and the control law are
derived and verified on both the physical arm and a matched simulation. Simulation is then used to study
how the same framework extends to other arm morphologies. Finally, the arm demonstrates tasks that have been difficult for both rigid
and soft arms: rejecting disturbances while writing on a moving whiteboard, and passively correcting
hidden misalignment during a key-insertion and drawer-opening task. That these tasks succeed under so straightforward a
controller is evidence for the advantage of this algorithm-informed structural design.

\end{abstract}


\section{Introduction}
\label{sec:intro}

\subsection{Motivation}

Robots that operate in unstructured and partially unobservable environments must do more than reach
accurate Cartesian positions. Tasks such as reaching into a bag, inserting a tool into a partially
occluded opening, or pulling a constrained object require the robot to maintain spatial precision while
remaining physically compliant to contact. In these settings, compliance is more than a safety
feature: it is a mechanism for passive error correction, letting the environment guide the
end-effector through small misalignments that are difficult to sense or model explicitly
\cite{whitney1982quasi}.

The compliance these tasks need is not a single fixed property. A manipulator may need to be soft in
one direction to absorb an unknown disturbance, stiff in another direction to preserve tool position,
and able to change this directional profile while it moves. Useful compliance is therefore controlled
and quantitatively adjustable, rather than a fixed material property or a binary, manually selected
mode. This motivates a robot that can control both tip position and quantitative task-space compliance
in real time.

\subsection{Literature Review}

Conventional rigid robots can emulate compliant behavior through impedance or admittance control
\cite{hogan1984impedance}. However, this ``compliance via control'' depends on sensing, controller
bandwidth, and actuator response. During fast contact or impact, the commanded behavior can lag the
physical interaction \cite{vanderborght2013variable}. Structurally soft robots instead provide true
elastic compliance, because their bodies and actuators deform directly under load \cite{rus2015design}.
However, many soft systems are not designed for simultaneous high-precision positioning and quantitative
compliance shaping \cite{thuruthel2018control}. Our goal is real-time quantitative compliance
together with position control.

Real-time quantitative compliance and position control falls within the impedance-control family
\cite{hogan1984impedance}, and a substantial body of work shapes endpoint stiffness for redundant
manipulators \cite{rice2018passive, ajoudani2015reduced, knevzevic2024cartesian}. On rigid arms this is
relatively direct, because independent joint motors expose a structured, efficiently controllable
stiffness map. For conventional structurally compliant robots using motors, such a map is typically obtained with
variable-stiffness actuators (VSAs) \cite{vanderborght2013variable}, which add a dedicated
stiffness-modulation mechanism such as antagonistic springs and a second motor at each joint
\cite{howard2011constraint}. This mechanism is essentially dead weight for positioning. When collocated with the joint, it enlarges the
joint's mass and inertia, and this penalty compounds along a serial chain because every proximal joint
must also accelerate the added mass of all distal modules. Even when the actuators are placed remotely and connected through belts, cables, or linkages, the added transmission still carries its own mass and complexity.
The result is a heavier arm with reduced
control precision and lower usable payload, a trade-off documented across variable-stiffness actuator designs \cite{vsapayload}.

For soft robots specifically, many methods can vary stiffness, including layer jamming
\cite{kim2013novel}, granular jamming \cite{brown2010universal}, auxetic and morphology-based structures
\cite{ma2025variable}, origami and folded structural-morphology mechanisms \cite{origamivsm}, and
fluid-driven stiffness modulation \cite{remy2023fluid}. These approaches create useful elastic behavior.
However, they typically require additional hardware and change stiffness nonlinearly with the control
input. This nonlinearity makes it difficult to smoothly tune the compliance profile while simultaneously
tracking position.
For example, layer jamming can substantially raise a link's bending stiffness~\cite{kim2013novel}, but under large loads the jammed layers can slip irreversibly, producing hysteresis and a stiffness that is not a smooth, repeatable function of the control input.

Pneumatic antagonistic stiffening is a notable method, because joint stiffness scales approximately
linearly with pressure. Bruder et al. developed a McKibben-actuated arm that combines variable stiffness
with posture control \cite{bruder2023scirob}. However, with a limited number of independently controlled
degrees of freedom, stiffness is regulated at the segment level through a single channel. There is
therefore not enough actuation to control compliance and position at the same time, because stiffening a
segment forces it straight in every direction. Decoupling compliance control from position control is therefore required.

Stella et al. demonstrated combined compliance and position control by co-optimizing shape and
segment-level stiffness \cite{stella2023prescribing}. Their system used tendons for posture control and a
pneumatically stiffened segment as the base.
Their stiffening is segment-level and
isotropic, so the stiffness profile available at a single posture is limited. In addition, the continuum
structure carries uncontrolled compliance that can buckle under heavy tendon load, which breaks the
piecewise-constant-curvature assumption \cite{webster2010design,continuum_buckling}.
This work relies on global
optimization to find different postures that realize a different compliance profile. Global optimization is too slow for real-time feedback control and does not enforce joint-space continuity along target trajectories.
Constantly switching to very different postures is undesirable
in practice, because real tasks often confine the robot to a specific posture in a tight workspace.

More generally, highly coupled continuum structures usually require global optimization to co-optimize
shape, pressure, and stiffness. Such optimization may be sufficient in static cases, but its computation
time and lack of joint-space continuity make it poorly suited to dynamic interaction, which involves
continuous changes in the desired compliance and position.

\subsection{Research Gap}

Building a capable real-time compliance and position controller is therefore a two-fold challenge. On
the hardware side, the design must balance system complexity, weight, and size against the number of
compliance-controllable degrees of freedom and the linearity of the input-to-compliance mapping.
On the modeling and
control side, real-time operation rules out heavy coupling: it requires a sparse joint-compliance
mapping in which the compliance of one joint does not influence another.

Impedance control with a compliance Jacobian already offers a promising route to real-time compliance
control \cite{rice2018passive}, but it is not readily applicable to existing soft hardware. Soft robots
are usually built with too much coupling to apply such algorithms directly: a single actuator typically deforms and stiffens an entire continuum segment at once, so the local compliance of one region cannot be set independently of its neighbors or of the arm's posture.
By contrast, rigid arms that have independent joint motors, such as
the UR5, the Franka Emika Panda, and the KUKA LBR iiwa, are far better suited to them. We argue instead for
algorithm-informed soft-robot design, in which the hardware is built specifically to satisfy the
requirements of the controller. Starting from the compliance-Jacobian requirement, the robot should
decouple joint angle from joint stiffness and provide as many such decoupled degrees of freedom as
possible.

Zuo et al.\ introduced a \emph{rigid-soft} arm \cite{zuo2025umarm} that builds a rigid articulated backbone into a soft, pneumatically actuated body. That work explored the possibility of densely packing many independently controlled actuators into a compact body, and demonstrated simultaneous compliance and position control. Its inverse-compliance behavior, however, was only partially explored: variable stiffness was demonstrated only at the static neutral pose, with every joint angle at zero and the arm hanging straight down, so a general methodology for commanding a target compliance at an arbitrary posture was missing. Beyond this missing methodology, several hardware limitations would themselves prevent effective compliance control. First, its structure cannot
withstand high-pressure antagonistic stiffening, because the off-the-shelf universal joints it uses are not
designed for such loads. Second, its closely packed pneumatic muscles physically rub against one another as they inflate.
This is tolerable for
closed-loop position control because feedback can correct position errors, but it breaks compliance modeling assumptions.
 Third, the actuator routing gives only a limited range of
motion, which constrains the robot's capability. Even so, antagonistic stiffening is a strong candidate for
real-time compliance and position control, because it decouples joint angle from joint stiffness and
adjusts joint stiffness linearly with pressure.

\subsection{Contributions}

This paper addresses the need for structurally soft robot hardware that is capable of simultaneous real-time position and compliance control, and offers methods to achieve such control. It meets this need with a rigid-soft robotic arm and a real-time controller for simultaneous
inverse kinematics (IK) and inverse compliance (IC). We use the term algorithm-informed hardware to
describe a design philosophy in which the mechanical structure is chosen to expose a sparse, locally
linearizable model for fast control, rather than leaving all stiffness coupling to be resolved by global
optimization.

The main contributions are:
\begin{itemize}
    \item \textbf{Algorithm-informed hardware design:} A hyper-redundant \cite{chirikjian1994modal}
    rigid-soft pneumatic arm whose redesigned joint mechanics decouple joint compliance sufficiently to
    bypass repeated global optimization and yield a rich tip-compliance profile at any static posture.
    \item \textbf{Real-time position and compliance control:} An iterative controller that combines
    standard Jacobian-based inverse kinematics (IK) with a compliance-Jacobian inverse compliance (IC)
    solver, null-space posture relief, and a dual-loop joint-angle/joint-energy pneumatic controller.
    The controller draws on impedance control and is tailored to the requirements of this hardware.
    \item \textbf{Generalized, reconfigurable modeling:} A morphology-general compliance and kinematics
    model.
    The modeling and control framework can accommodate arbitrary segment lengths and inter-segment connector orientations,
    which lets the same arm be reconfigured for
    task-appropriate compliance.
\end{itemize}

We first evaluate the controller through algorithmic benchmarks that compare the iterative solver
against a global optimization baseline in static and dynamic compliance-tracking tasks. We then validate
the analytical compliance model against both real-system displacement-to-force measurements and simulated
force-to-displacement measurements, followed by a direct posture comparison in which the simulated and
physical systems receive the same pressure target. This validation supports using the simulation to
examine morphology changes that are not yet practical to manufacture. One such change is a reconfiguration
that removes the baseline compliance singularity and reshapes the attainable compliance workspace,
demonstrating that the same modeling and control generalize across morphologies. Finally, we demonstrate the
hardware in tasks that require both precision and passive compliance: writing on a moving whiteboard and
correcting unobservable misalignment during a key insertion and drawer-opening task.

\section{Methods}
\label{sec:methods}

\subsection{Design}

The hardware is designed around the requirements of the real-time controller. First, we build a rigid articulated backbone into a joint-wise soft, pneumatically actuated
body. Second, we densely install many actuators in compact segments. Third, we mechanically isolate the actuators of each joint from those of its neighbors. This exposes
a joint-space stiffness map that is sparse, because each joint's stiffness is set almost independently of
the others. The stiffness map is also linear in the actuator pressures, so it is cheap to differentiate.
This is precisely what lets the controller treat compliance with the same differential machinery it uses for
position.

\begin{figure}[t]
    \centering
    \includegraphics[width=\linewidth]{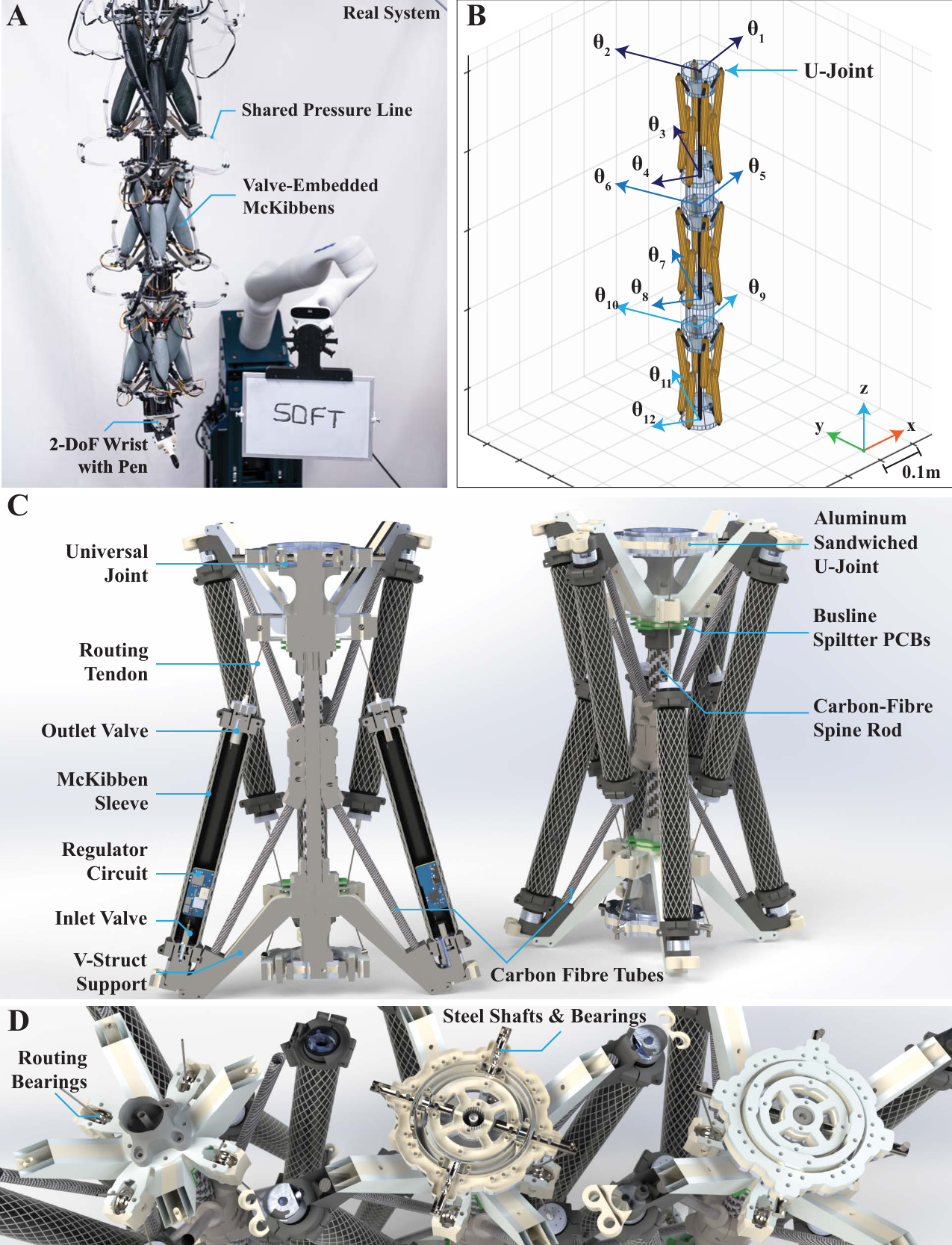}
    \caption{(A) Photo of the three-segment rigid-soft arm. (B) Skeleton model at the
        $\mathbf{q}=\mathbf{0}$ configuration. (C) Antagonistic McKibben actuator pair routed through
        tendons and bearings to rotate a revolute joint. (D) Custom U-joint structure with outer,
        middle, and inner rings for tendon attachment, bearing support, and link connection.}
    \label{fig:arm_structure}
\end{figure}

The manipulator is a 7-link rigid-soft chain with 6 custom universal joints (U-joints) grouped into three segments, producing 12
revolute joint coordinates collected in
$\mathbf{q} \in \mathbb{R}^{12}$. At $\mathbf{q}=\mathbf{0}$, the arm hangs downward in its nominal
resting configuration. Each U-joint contains two perpendicular revolute axes, and each axis is driven
by an antagonistic pair of McKibben actuators~\cite{tondu2012modelling}. Thus, each U-joint is driven by four independently
regulated pneumatic muscles, for a total of 24 actuators.

The U-joints were manufactured from water-jet aluminum plates and nylon printed spacers. The cascaded
ring layout shown in Fig.~\ref{fig:arm_structure} assigns a role to each ring: the center link connects to
the inner ring, the bearings sit in the middle ring, and the tendon attachments load the outer ring. Compared
with off-the-shelf U-joints, this geometry was chosen for high antagonistic load capacity, low slack,
and low mass. The V-shaped actuator supports route longer McKibben actuators around the joint while
offsetting neighboring muscles to reduce interference; braided carbon-fiber rods reinforce these
supports with minimal added mass. Routing the tendons this way also lets each U-joint move through
tightly confined spaces, giving a larger joint range of motion than a similar design found in \cite{zuo2025umarm}.

This layout intentionally incorporates a rigid structural backbone into a soft-actuated system. Actuators that
control one U-joint are mechanically isolated from actuators controlling neighboring U-joints. Within a
single U-joint, the two perpendicular axes can exhibit mild coupling at large joint angles, but the
coupling remains local to that U-joint. Complete mechanical decoupling of all 12
revolute axes would require substantially more hardware and would reduce actuator density. The chosen
architecture therefore preserves the algorithmically important structure: joint stiffness is locally
coupled inside each U-joint but decoupled across U-joints.

The McKibben actuators drive the arm and modulate joint compliance. Following the energy formulation in
\cite{bruder2018force}, the internal energy of a single actuator is
\begin{equation}
U_{\mathrm{act}} = v_{\mathrm{act}} p_{\mathrm{act}}
= \frac{B^2-L^2}{4\pi N^2} L p_{\mathrm{act}},
\label{eq:actuator_energy}
\end{equation}
where $v_{\mathrm{act}}$ is the actuator volume, $B$ is the fiber length, $N$ is the number of fiber revolutions around the actuator, $L$ is the
current actuator length, and $p_{\mathrm{act}}$ is the internal pressure. The negative gradient of this energy
with respect to actuator length gives the pressure-length force relationship
\begin{equation}
F_{\mathrm{act}}(L,p_{\mathrm{act}})
= -\frac{\partial U_{\mathrm{act}}}{\partial L}
= -\frac{B^2-3L^2}{4\pi N^2}p_{\mathrm{act}} .
\label{eq:actuator_force}
\end{equation}
The total actuator potential energy $U_{\mathrm{total}}$ is the sum of
Eq.~\ref{eq:actuator_energy} over all actuators.

Each McKibben actuator integrates its own pressure-regulating valve and local controller, a valve-embedded design \cite{zuo2024embedded}.
Embedding pressure regulation inside each actuator reduces tubing and pressure delay compared
with centralized regulators.
Our system uses a CAN bus for communication, which provides real-time pressure feedback from the
distributed controllers and supports higher-frequency command updates. We have verified a stable
150\,Hz control and feedback rate for all 24 actuators on the same bus.
A high-flow pressure controller is used on the topmost segment, which experiences the most load and
disturbance from the lower segments.

\begin{figure}[t]
    \centering
    \includegraphics[width=\linewidth]{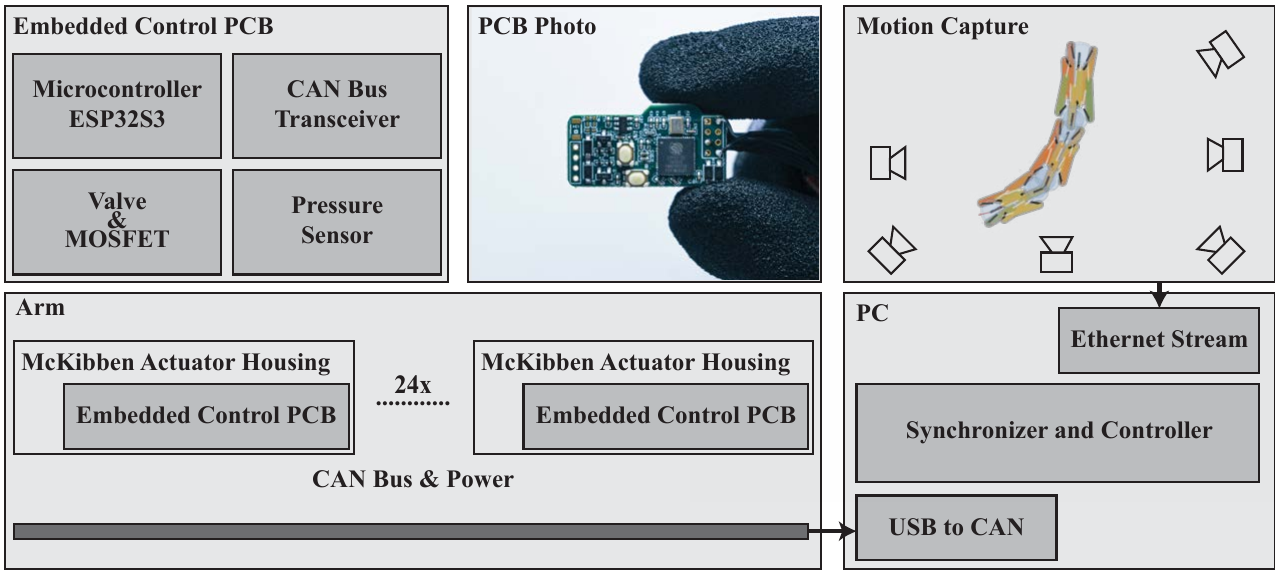}
    \caption{Hardware system schematic. A PC-side controller computes the real-time IK/IC solution and
        sends pressure and joint-energy commands over a CAN bus to the distributed valve-embedded
        controllers, which regulate the 24 McKibben actuators driving the six U-joints and stream
        measured pressure back over the same bus. An external motion-capture system tracks markers on the
        links to reconstruct joint angles and tip pose for command tracking and validation.}
    \label{fig:hardware_schematic}
\end{figure}

Fig.~\ref{fig:hardware_schematic} shows how these pieces connect into a single real-time loop. The
PC runs the controller software. The CAN bus carries pressure and joint-energy commands out
to the valve-embedded controllers and measured pressures back. The 24 McKibben actuators apply the
resulting joint torques across the six U-joints. Finally, an external motion-capture system closes the loop
with link-pose feedback that reconstructs joint angles and tip pose for control.

These hardware features are essential for deploying controllers on the real system. Keeping
pressure regulation, sensing, and communication distributed and local to each actuator helps the system
run fast enough to be controlled in real time. At the same time, ensuring that each actuator's actual
pressure tracks its target pressure closely enough helps achieve precise compliance shaping and position
control.

\subsection{Compliance Modeling}

To model task-space compliance, we use an energy-based formulation adapted from Bruder et al.
\cite{bruder2023scirob}. The generalized joint torques are
computed from the negative gradient of total actuator energy with respect to the joint configuration,
\begin{equation}
\boldsymbol{\tau}
= -\frac{\partial U_{\mathrm{total}}(\mathbf{q},\mathbf{p})}{\partial \mathbf{q}},
\label{eq:joint_torque_energy}
\end{equation}
where $\mathbf{p}$ collects actuator pressures. The joint-space stiffness matrix is the differential
change in these torques under a virtual joint displacement,
\begin{equation}
K_{\mathbf{q}}(\mathbf{q},\mathbf{p})
= -\frac{\partial^2 U_{\mathrm{total}}(\mathbf{q},\mathbf{p})}{\partial \mathbf{q}^2}.
\label{eq:joint_stiffness_energy}
\end{equation}

Because neighboring U-joints are mechanically isolated, this stiffness matrix has a sparse block
structure. For a fixed configuration $\mathbf{q}$, the $12 \times 12$ joint-space stiffness matrix can
be written as
\begin{equation}
K_{\mathbf{q}} =
\begin{bmatrix}
K_{\mathrm{U},1} & & \mathbf{0} \\
& \ddots & \\
\mathbf{0} & & K_{\mathrm{U},6}
\end{bmatrix},
\quad
K_{\mathrm{U},j} =
\begin{bmatrix}
k_{j,1} & k_{j,c} \\
k_{j,c} & k_{j,2}
\end{bmatrix}.
\label{eq:block_stiffness}
\end{equation}
The diagonal entries $k_{j,1}$ and $k_{j,2}$ describe stiffness about the two revolute axes of the
$j$th U-joint, and $k_{j,c}$ captures the local coupling between those axes. The off-diagonal blocks
between different U-joints are zero.

The same structure also makes the pressure-to-stiffness map efficient. Since actuator energy is linear
in pressure, and stiffness is obtained by differentiating that energy with respect to configuration,
the independent entries of each U-joint stiffness block vary linearly with the four pressures driving
that U-joint. We write
\begin{equation}
\mathbf{k}_j =
\begin{bmatrix}
k_{j,1} & k_{j,2} & k_{j,c}
\end{bmatrix}^{\intercal}
= G_j(\mathbf{q})\mathbf{p}_j,
\quad
G_j(\mathbf{q}) \in \mathbb{R}^{3 \times 4},
\label{eq:pressure_stiffness_map}
\end{equation}
where $\mathbf{p}_j$ contains the four actuator pressures associated with the $j$th U-joint. The entries of $G_j(\mathbf{q})$ are derived analytically from the actuator lengths, tendon moment arms,
and McKibben force model; these expressions follow directly from the actuator energy model in Eq.~\ref{eq:actuator_energy} and the joint geometry.
For a fixed $\mathbf{q}$, $G_j(\mathbf{q})$ is constant, so the block entries of $K_{\mathbf{q}}$ can be assembled rapidly without expensive differentiation.

The joint-space compliance matrix is the inverse of the stiffness matrix, which is non-singular provided all joint stiffnesses are non-zero,
\begin{equation}
C_{\mathbf{q}}(\mathbf{q},\mathbf{p}) = K_{\mathbf{q}}^{-1}(\mathbf{q},\mathbf{p}).
\label{eq:joint_compliance}
\end{equation}
Using the manipulator Jacobian $J(\mathbf{q})$~\cite{lynch2017modern}, the task-space compliance at the end-effector is
computed by the transformation
\begin{equation}
C_{\mathbf{x}}(\mathbf{q},\mathbf{p})
= J(\mathbf{q}) C_{\mathbf{q}}(\mathbf{q},\mathbf{p}) J^{\intercal}(\mathbf{q}).
\label{eq:task_compliance}
\end{equation}
This matrix provides a quantitative local description of directional tip compliance for a given
configuration and pressure state.

\subsection{Real-Time Iterative Controller}

To control tip compliance and position in real time, we must answer a key question: which control
inputs produce a differential change in compliance, and which produce a differential change in position?
Answering it lets us continuously command the robot along a smooth compliance and position trajectory.

For position control, we model the forward kinematics with the standard product-of-exponentials method.
For inverse kinematics, given a desired tip trajectory $\mathbf{x}_d(t)$, the manipulator Jacobian
$J(\mathbf{q})$ maps a joint update to a local end-effector update. We solve it with damped least squares
(DLS) for numerical stability near singularities, together with a boundary function that repels the
joints from their physical limits. A null-space projector is also available for secondary goals that do
not disturb position control. This preserves the familiar structure of redundant rigid-robot IK \cite{lynch2017modern, buss2004introduction} while
using soft pneumatic actuators as the physical drive system.

For compliance control, we construct a compliance Jacobian that relates small changes in the actuator pressures to small changes in the end-effector's task-space compliance, in the same way the manipulator Jacobian relates joint-angle changes to tip motion. The task-space compliance matrix $C_{\mathbf{x}}$, defined in Eq.~\ref{eq:task_compliance}, quantifies how far the tip deflects under a unit force applied in each direction. The compliance Jacobian then lets the controller steer this compliance toward a desired profile, by taking small, feasible steps in joint stiffness that are realized through changes in the antagonistic pressures. Several prior works have used derivatives
of task-space compliance or stiffness matrices for differential stiffness shaping on rigid or
variable-stiffness robots \cite{ajoudani2015reduced,rice2018passive,knevzevic2024cartesian,howard2011constraint}.
Here, the same differential idea is adapted to the pressure-dependent stiffness map of a
hyper-redundant soft-actuated arm. Let $\mathbf{u}$ denote the active compliance-control variables,
chosen as the leading pressures of the antagonistic actuator pairs. Each revolute axis is driven by an antagonistic pair of actuators. Holding a given joint angle requires the two pressures to satisfy a static torque-equilibrium relationship at that joint. One pressure of each pair is therefore designated the leading pressure and used as the independent control variable. The other pressure, called the trailing pressure, is then fixed directly by that relationship. Because the joint torque is affine in the two pressures, this relationship is affine, so the trailing pressure is a linear function of the leading pressure; substituting it into Eq.~\ref{eq:pressure_stiffness_map} reduces each U-joint stiffness block to a function of its two leading pressures alone, with an effective map $\tilde{G}_j(\mathbf{q})\in\mathbb{R}^{3\times2}$. Each antagonistic pair thus contributes a single compliance-control variable while preserving the pressure relationship needed to hold the joint angle.

For a fixed configuration, $J(\mathbf{q})$ is treated as constant while differentiating the task-space
compliance matrix. The sensitivity of $C_{\mathbf{x}}$ with respect to the $i$th control variable is
\begin{equation}
\frac{\partial C_{\mathbf{x}}}{\partial u_i}
= J(\mathbf{q})
\frac{\partial C_{\mathbf{q}}}{\partial u_i}
J^{\intercal}(\mathbf{q}).
\label{eq:task_compliance_derivative}
\end{equation}
Using the matrix-inverse derivative identity~\cite{petersen2012matrix} of Eq.~\ref{eq:joint_compliance},
\begin{equation}
\frac{\partial C_{\mathbf{q}}}{\partial u_i}
= -C_{\mathbf{q}}
\frac{\partial K_{\mathbf{q}}}{\partial u_i}
C_{\mathbf{q}}.
\label{eq:joint_compliance_derivative}
\end{equation}
The remaining stiffness derivative is sparse. If $u_i$ corresponds to the $a$th leading pressure of
the $j$th U-joint, and $\mathbf{g}_{j,a}(\mathbf{q})=[g_{j,a}^{(1)},g_{j,a}^{(2)},g_{j,a}^{(c)}]^{\intercal}$
is the corresponding column of the effective map $\tilde{G}_j(\mathbf{q})$, then
\begin{equation}
\frac{\partial K_{\mathbf{q}}}{\partial u_i}
= \operatorname{blkdiag}\left(\mathbf{0},\ldots,
\begin{bmatrix}
g_{j,a}^{(1)} & g_{j,a}^{(c)} \\
g_{j,a}^{(c)} & g_{j,a}^{(2)}
\end{bmatrix},
\ldots,\mathbf{0}\right).
\label{eq:stiffness_pressure_derivative}
\end{equation}

Because $C_{\mathbf{x}}$ is symmetric, it has six independent entries, which we vectorize as
$\operatorname{vec}(C_{\mathbf{x}})=[C_{00},C_{11},C_{22},C_{10},C_{20},C_{21}]^{\intercal}$.
Applying this
vectorization to the sensitivity in Eq.~\ref{eq:task_compliance_derivative}, the $i$th column of the
compliance Jacobian is $[J_c]_{:,i}=\operatorname{vec}(\partial C_{\mathbf{x}}/\partial u_i)$ for
$i=1,\dots,12$. The resulting matrix $J_c\in\mathbb{R}^{6\times12}$ maps the twelve leading-pressure updates
to the six independent compliance entries,
\begin{equation}
\Delta \operatorname{vec}(C_{\mathbf{x}}) \approx J_c \Delta\mathbf{u}.
\label{eq:compliance_jacobian_map}
\end{equation}
Given the compliance error $e_c=\operatorname{vec}(C_{\mathbf{x},d}-C_{\mathbf{x}})$, the primary update
is computed with a bounded damped least squares (DLS) solve,
\begin{multline}
\left(J_c^{\intercal}J_c + \lambda^2 I + \mu H_B\right)
\Delta \mathbf{u}_{\mathrm{primary}} \\
= J_c^{\intercal}e_c - \mu \nabla B .
\label{eq:bounded_compliance_solve}
\end{multline}
Here $\lambda$ is a damping factor, $B$ is an interior-point pressure-boundary function, $H_B$ is its
local curvature term, and $\mu$ weights the boundary penalty. The damping term prevents unstable
updates near singular compliance configurations, while the boundary term reduces pressure-limit
chatter caused by repeatedly clipping infeasible commands.
Concretely, the update is a damped least-squares step on the compliance error \cite{buss2004introduction} augmented with an interior-point log-barrier penalty that enforces the pressure limits \cite{boyd2004convex}. The barrier is
\begin{equation}
B(\mathbf{u}) = -\sum_i \left[\ln\!\left(u_i - u_i^{\min}\right) + \ln\!\left(u_i^{\max} - u_i\right)\right],
\label{eq:log_barrier}
\end{equation}
where $[u_i^{\min},u_i^{\max}]$ bounds the $i$th leading pressure; its gradient $\nabla B$ and Hessian $H_B=\nabla^2 B$ grow without bound as any pressure approaches a limit. In Eq.~\ref{eq:bounded_compliance_solve}, $J_c^{\intercal}J_c+\lambda^2 I$ is the standard damped Gauss-Newton term for the compliance residual, while $\mu H_B$ and $\mu\nabla B$ add the barrier's curvature and gradient so each update is pulled back inside the feasible pressure box.

The arm has more compliance-control variables than the six independent entries of
$C_{\mathbf{x}}$, so the compliance Jacobian has a non-trivial null space. We use the projector
$N_c=I-J_c^+J_c$ to add secondary objectives that do not alter the first-order compliance update. One
secondary objective minimizes stored actuator energy. A second coupling between inverse kinematics (IK) and inverse compliance (IC) uses joint
energy as a posture signal. When a joint requires high co-contraction energy to reach the target
compliance, the IK null-space bias can move that joint toward a more neutral posture that has a larger
feasible stiffness workspace. This bias preserves the end-effector position.

On hardware, direct pressure commands from the solver are not sufficiently robust because the ideal
pairwise pressure relationship is never exact. Instead, the high-level IC layer converts the pressure
update into desired joint energy, and the low-level controller tracks desired joint energy and desired
joint angle simultaneously using proportional-integral-derivative (PID) loops. The angle loop regulates differential pressure across an antagonistic pair,
while the energy loop regulates antagonistic pressure. This gives the controller a practical bridge
from model-based compliance updates to physical pneumatic actuation.

\begin{algorithm}[t]
    \caption{Real-time inverse kinematics, inverse compliance, and hardware execution loop}
    \label{alg:dual_loop_pid}
    \begin{algorithmic}[1]
        \REQUIRE Desired tip trajectory $\mathbf{x}_d(t)$, desired compliance
        $C_{\mathbf{x},d}(t)$, robot model, joint limits, and pressure limits
        \ENSURE Hardware pressure commands $\mathbf{p}_{\mathrm{cmd}}$
        \STATE Initialize $\mathbf{q}$, $\mathbf{p}$, controller gains, filters, and safety limits
        \WHILE{real-time controller is enabled}
            \STATE Read joint angles $\mathbf{q}$ and actuator pressures $\mathbf{p}$
            \STATE Compute $\mathbf{x}(\mathbf{q})$ and $C_{\mathbf{x}}(\mathbf{q},\mathbf{p})$
            \STATE Compute IK error $e_x=\mathbf{x}_d-\mathbf{x}(\mathbf{q})$
            \STATE Update $\mathbf{q}_d$ with DLS IK and null-space posture terms
            \STATE Compute compliance error $e_c=\operatorname{vec}(C_{\mathbf{x},d}-C_{\mathbf{x}})$
            \STATE Build $J_c$ at the current state and solve Eq.~\ref{eq:bounded_compliance_solve}
            \STATE Convert the compliance update to desired joint energies $\mathbf{E}_d$
            \FOR{each antagonistic actuator pair $r$}
                \STATE Compute angle error $e_{q,r}=q_{d,r}-q_r$ and energy error
                $e_{E,r}=E_{d,r}-E_r$
                \STATE PID$_q(e_{q,r})$ updates differential pressure $\Delta p_r$
                \STATE PID$_E(e_{E,r})$ updates antagonistic pressure $\bar{p}_r$
                \STATE Map $(\bar{p}_r,\Delta p_r)$ to bounded actuator commands
                $(p_{r,+},p_{r,-})$
            \ENDFOR
            \STATE Send $\mathbf{p}_{\mathrm{cmd}}$ to the pneumatic controllers over CAN bus
        \ENDWHILE
    \end{algorithmic}
\end{algorithm}
\section{Algorithmic Benchmarking}
\label{sec:algorithmic_benchmarking}

The proposed iterative IC solver is a \emph{local} method: at each control cycle it linearizes the
compliance map and takes one damped step, converging to a local rather than a global minimum of the
compliance error. A global optimizer can in principle reach a lower final error. However, it must finish
searching before any pressure command is available, so it cannot respond within a control cycle. It also
solves each target independently, so its commands may not be continuous between successive targets. Our
claim is that the arm is redundant enough that the local solution suffices in practice. This
redundancy comes both from the joint-space null space for position and from the over-parametrized actuator
pressures for compliance. The local solution is also fast enough to run alongside the moving robot, and it
is continuous in joint space by construction. The
compliance-Jacobian formulation for combined inverse kinematics and compliance is, in its basic form, a
local solver. Our controller is essentially this established local method, adapted with the changes needed
to run on a structurally soft, pneumatically actuated system whose hardware was designed to suit it, rather
than a fundamentally new algorithm. Rice et al. extend the same formulation into a globally optimal solver
by pairing it with homotopy methods~\cite{rice2020homotopy}; we mention this route for completeness but do
not pursue it here.
We test this in two stages.
First, a static profile-to-profile benchmark gives the global optimizer its most favorable setting, in
which only the final state matters and waiting for a solution is acceptable. For this baseline we use
differential evolution (DE) as a representative derivative-free search over feasible actuator pressures. Second, a figure-8 shape-tracking benchmark with a changing compliance target exposes the
global method under continuous motion and compliance switching, where its computation delay and command
discontinuity rule it out.

\subsection{Static Pose-to-Pose Reaching}

We benchmark the two solvers on 200 randomized profile-to-profile transitions. For each case we sample
two gravity-equilibrium actuator-pressure sets. Each set is a vector of leading actuator pressures for which the arm rests in static equilibrium under gravity, so that the actuator torques exactly balance the gravitational load at the corresponding joint configuration.
 The first defines the target: its configuration fixes
the target tip position $\mathbf{x}^*$ and the target task-space compliance $C_{\mathbf{x}}^*$, and that
generating configuration is never revealed to the solvers. The second defines the initial pose and
pressure. The global baseline first runs weighted-least-norm IK from the initial pose to reach
$\mathbf{x}^*$, then runs DE over the leading actuator pressures to match $C_{\mathbf{x}}^*$ at the
reached configuration. At each of up to 250 iterations, the proposed solver instead takes one IK step toward $\mathbf{x}^*$
together with one inverse-compliance step. This inverse-compliance step is a bounded damped-least-squares
update with a log-barrier feasibility term and a null-space term that minimizes McKibben pneumatic energy. Both solvers are evaluated analytically rather than in simulation or on hardware: for each solver, we
pair the actuator pressures with the joint configuration that satisfies static torque equilibrium under
gravity.

Because both pipelines reach $\mathbf{x}^*$ with the same IK, their position error is identical
($0.160$\,mm mean), and the comparison reduces to compliance accuracy, speed, and energy.
Table~\ref{tab:inverse_compliance_benchmark} reports the per-metric means over the 200 transitions. As
expected, the global optimizer is the more accurate compliance solver. It reaches a $1.13\%$ mean
relative compliance error against $6.54\%$ for the iterative method. This gap arises because the global
optimizer searches the pressure space for the global match, while the local solver settles at the nearest
feasible minimum. That gap is
real but modest, and it buys the local method two operational advantages. First, the iterative solver is
about $5.4\times$ faster ($56$\,ms versus $301$\,ms mean), fast enough to emit a usable command within a
control cycle rather than only after a one-shot search completes. Second, its null-space term drives it
to a lower-energy solution ($343.9$\,J versus $406.5$\,J mean, roughly $15\%$ less), which is cheaper to
hold pneumatically. The iterative method also advances the joint configuration in continuous steps, so
successive commands are joint-continuous by construction. The global solver, in contrast, re-optimizes each
target independently and would need an additional, costly continuity constraint to match this.

This is the outcome the redundancy argument predicts. The arm has more actuated degrees
of freedom than independent compliance control parameters, and it also retains a joint-space null space for
position. As a result, many configurations satisfy a given target closely. A local solver that stops at the nearest such
configuration therefore sacrifices little accuracy while gaining speed, lower energy, and continuity. This
trade-off favors real-time use even in this static setting. Note that this benchmarking
static setting is the global optimizer's best case.

\begin{table}[t]
    \centering
    \caption{Inverse-compliance solver benchmark: per-metric means over 200 transitions.}
    \label{tab:inverse_compliance_benchmark}
    \begin{tabular}{lcc}
        \hline
        Metric & Global (DE) & Iterative (ours) \\
        \hline
        Compliance error (\% rel.) & 1.13 & 6.54 \\
        Solve time (ms) & 301.1 & 55.6 \\
        System energy (J) & 406.5 & 343.9 \\
        \hline
    \end{tabular}
\end{table}

\begin{figure}[t]
    \centering
    \includegraphics[width=\linewidth]{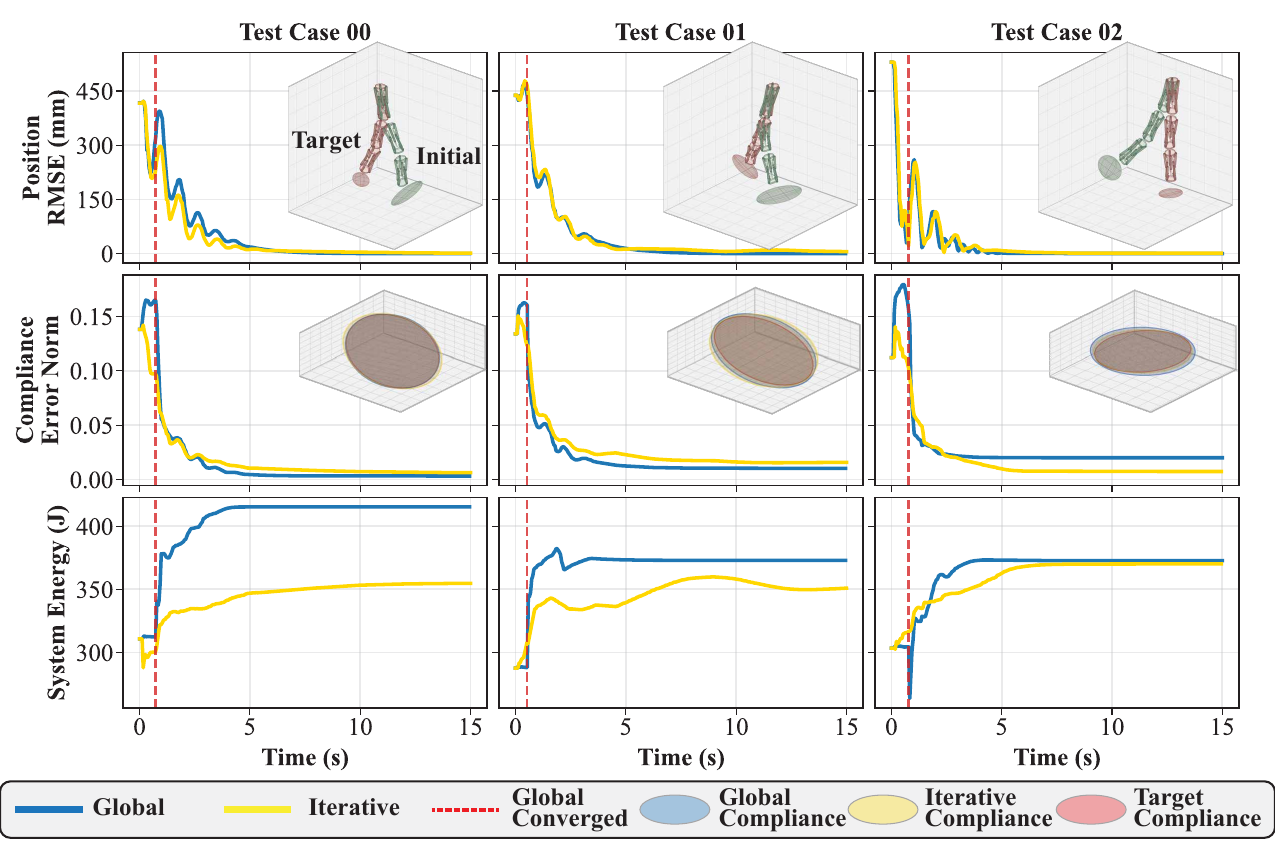}
    \caption{Static two-pose inverse-compliance benchmark. Columns show three fixed-pose test cases.
        (A) Position RMSE during the transition; insets show the target and initial robot postures.
        (B) Compliance error norm and final compliance ellipses for the target, global optimizer, and
        compliance-Jacobian iterative controller. (C) System energy. The red dash-dotted vertical line
        marks when the global optimizer finishes computing and its command becomes available.}
    \label{fig:compliance_2profile_test}
\end{figure}

Beyond the aggregate means, Fig.~\ref{fig:compliance_2profile_test} shows three representative two-pose
transitions. In each test case, the initial and target postures are fixed,
the target compliance profile is fixed, and the optimizer is allowed enough time to converge before
its final command is evaluated. Under these conditions, the global baseline can reach a final pose and
compliance ellipse close to the target and close to the iterative result. This is the appropriate use
case for global optimization: a static endpoint problem in which waiting for the final answer is
acceptable.

The same trial also shows why final accuracy alone is not sufficient for real-time compliance control.
The red dash-dotted line marks the time at which the global optimizer finishes computing and its
pressure command becomes available. Before that time, the iterative controller has already begun
reducing the compliance error. After both methods have converged, their final compliance ellipses are
similar. The compliance-Jacobian trajectory, however, reaches the target with substantially lower stored
system energy. Thus, even in the setting that most favors the global optimizer, the proposed method preserves the
main advantages needed for physical execution: immediate response and lower energy use.

\subsection{Dynamic Trajectory with Compliance Switching}

The dynamic benchmark tests the coupled IK/IC controller on a moving target. Unlike the analytical
static benchmark, it runs in the MuJoCo physics simulator with the full control loop closed. Here the
inverse-compliance solver's joint-angle and joint-energy targets are realized through the actual
dual-loop PID controller of Algorithm~\ref{alg:dual_loop_pid}. In that controller, an angle loop regulates
the differential pressure and an energy loop regulates the antagonistic pressure of each actuator pair.
The robot tracks a continuous figure-8 trajectory for three cycles while alternating between two
directional compliance profiles every half-loop: one profile is soft in $x$ and stiff in $y$, while the
other is stiff in $x$ and soft in $y$. This task is difficult for global optimization because both the Cartesian target
and the compliance target change continuously. Re-solving a full pressure optimization at each point
would introduce delay and can create discontinuous pressure commands between adjacent time steps.

\begin{figure}[t]
    \centering
    \includegraphics[width=\linewidth]{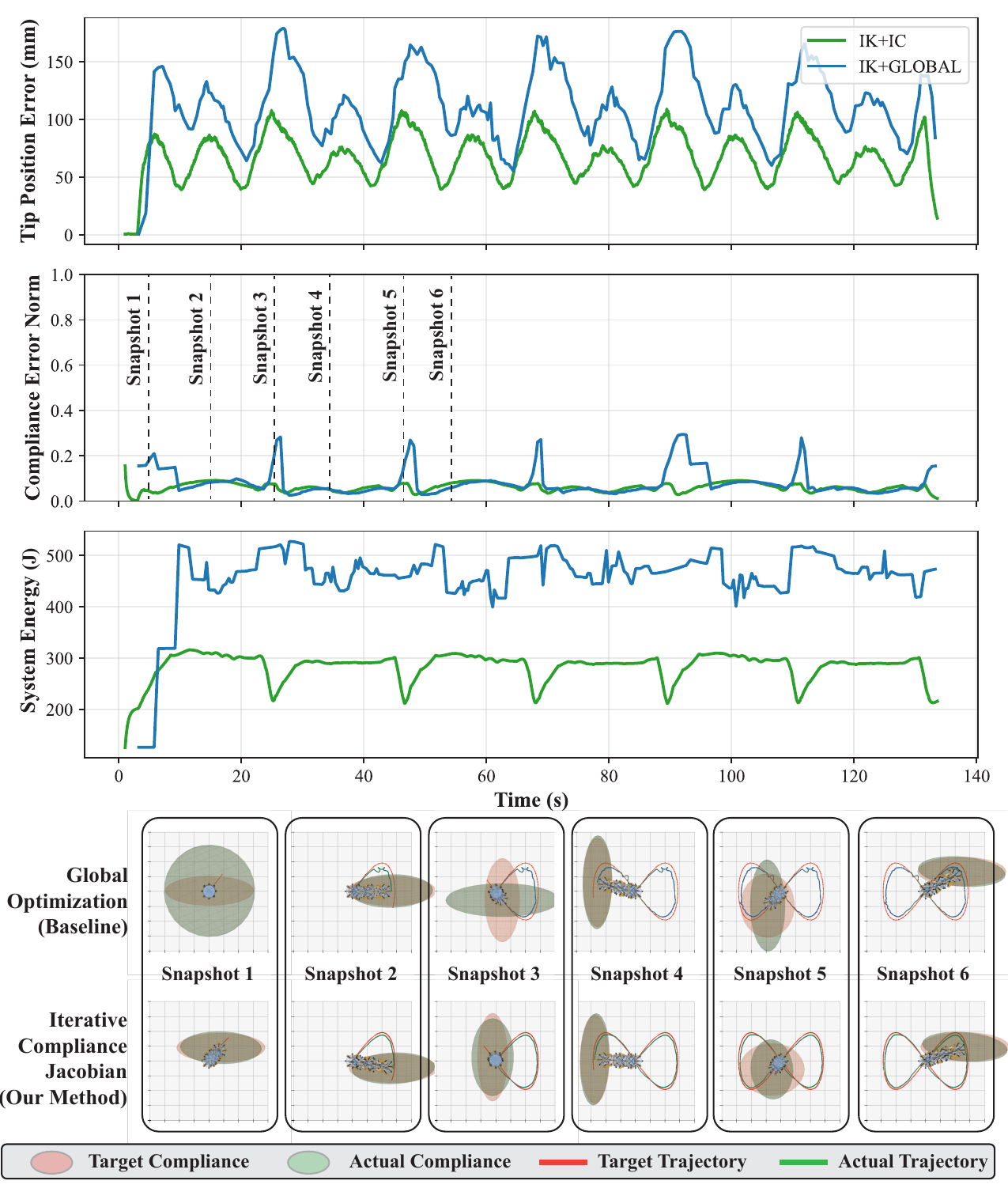}
    \caption{Dynamic figure-8 benchmark with alternating directional compliance targets. The robot
        tracks three figure-8 cycles while switching between $x$-soft/$y$-stiff and
        $x$-stiff/$y$-soft profiles every half-loop. (A) Position error. (B) Compliance error.
        (C) System energy. (D) Trajectory snapshots comparing the global optimization baseline and the
        compliance-Jacobian iterative controller. The dashed vertical lines (Snapshot~1--6) mark the time
        points at which the trajectory snapshots in (D) are taken.}
    \label{fig:compliance_draw8_test}
\end{figure}

The proposed controller executes the trajectory as one continuous differential process. The IK layer
keeps the end-effector on the figure-8 path, while the IC layer updates the pressure and energy targets
needed to switch compliance direction. Fig.~\ref{fig:compliance_draw8_test} compares position error,
compliance error, and system energy for the global optimization baseline and the compliance-Jacobian
iterative controller, with trajectory snapshots shown below the time histories. The iterative method
maintains smoother behavior through the compliance switches. Over the tracking phase, its tip-tracking
position RMSE is $70.5$\,mm, against $120.2$\,mm for the global baseline. This residual error is expected: the low-level loops use PID without feedforward, so they act on measured error alone and cannot drive it fully to zero. Its task-space compliance
error is also lower ($0.064$ versus $0.106$), and its average system energy is lower as well ($287$
versus $462$\,J). The iterative controller updates compliance at roughly $29$\,Hz, while the global
optimizer updates at about $1.3$\,Hz. This higher update rate keeps the controller locked to the
moving target, which contributes to the smoother tracking.

When only the final state matters, the
global optimizer can wait, compute, and then dispatch a good endpoint command. During the figure-8
task, however, the endpoint and compliance target have already moved by the time a delayed global
solution becomes available. The compliance-Jacobian controller avoids this wait-and-jump behavior by
updating pressure and energy targets continuously as the robot moves. Taken together, the two benchmarks
make the case directly. The local solver concedes only a modest amount of static compliance accuracy,
yet it is the only one of the two that keeps the robot moving in real time.

\section[Simulation and Model Validation]{Simulation and Model Validation}
\label{sec:model_validation_design}

The algorithmic benchmarks above test whether the proposed controller can update compliance quickly
and smoothly. We next ask whether the analytical compliance model and the
physics simulation describe the physical arm well enough to be trusted. This matters because several of
the designs we study later cannot yet be built and can only be evaluated in simulation. Before relying on
the simulation for that purpose, this section establishes up front that it agrees with the real hardware
on both compliance and pressure-driven posture. Once that agreement is shown, the simulation can serve as
a faithful guide to the real system's compliance behavior. The
validation proceeds in three complementary steps. First, the real system is probed with prescribed
tip displacements while measured reaction forces are mapped back through the analytical compliance
matrix. Second, the simulated system in MuJoCo is tested with the complementary force-controlled protocol,
where applied tip forces are compared against the resulting simulated displacements. Third, the
simulated system and the real system are compared directly under the same pressure target.

\subsection{Real-System Compliance Validation}

\begin{figure}[t]
    \centering
    \includegraphics[width=\linewidth]{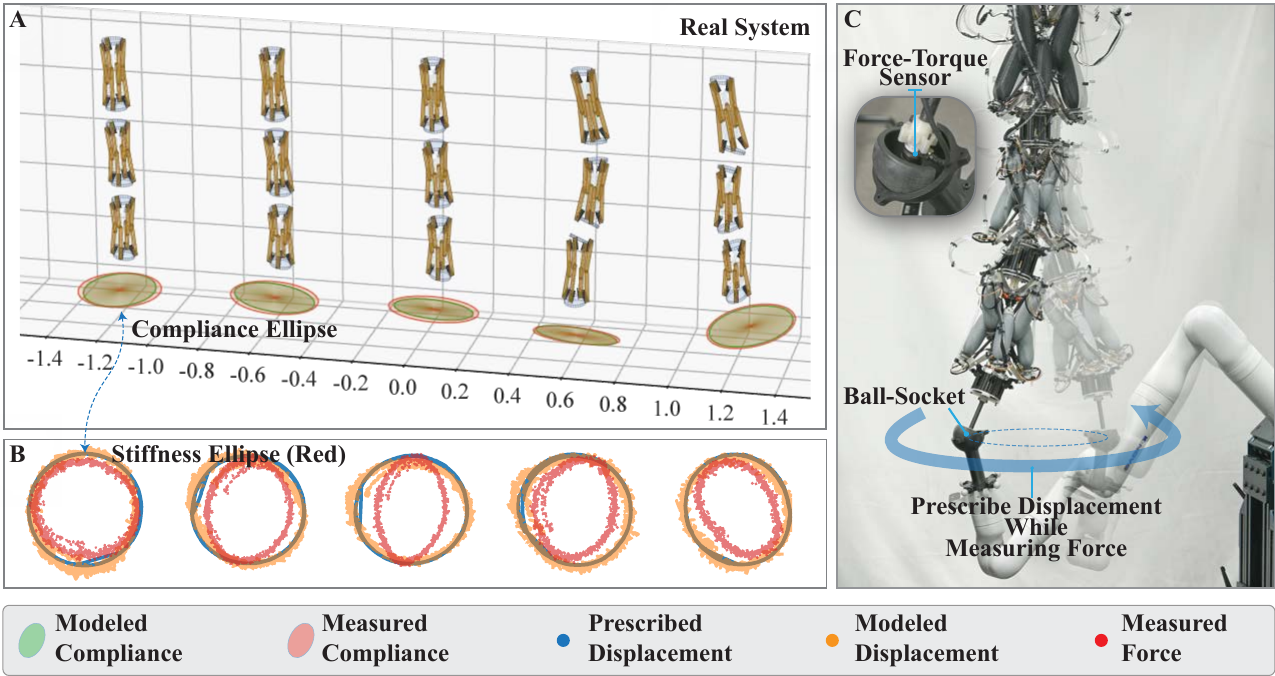}
    \caption{Real-system compliance probing and model validation. (A) Initial hardware
        configurations are shown with their corresponding endpoint compliance profiles; each
        compliance ellipse is plotted at the arm tip. (B) A circular tip displacement is prescribed (blue); the
        force-torque sensor then measures the reaction force generated by the perturbation (red), which
        traces a stiffness ellipse whose major axis is perpendicular to the high-compliance direction.
        Because the experiment is displacement-controlled, compliance is not measured directly: the
        measured force is instead mapped through the analytical compliance matrix $C_{\mathbf{x}}$ to give
        the modeled displacement (orange), which is compared against the prescribed displacement. The
        traces, in the order they are produced, are the prescribed displacement, the measured force,
        then the modeled displacement. Because the probe yields force--displacement pairs rather than a
        compliance matrix, the arm's empirical compliance ellipse is obtained by fitting, with a global
        optimizer, the matrix that best maps the measured forces to the prescribed displacements. (C) Experimental setup: a Kinova Gen3 arm connects to
        the bottom of the rigid-soft arm structure through a ball-socket joint to prescribe horizontal
        circular motion, while a force-torque sensor integrated inside the tip ball records the
        reaction forces.
        }
    \label{fig:real_system_compliance_probing}
\end{figure}

The real-system compliance test is displacement controlled because it is easier to prescribe small
repeatable tip motions on the physical robot than to apply repeatable tip forces. As shown in
Fig.~\ref{fig:real_system_compliance_probing}, the tested configurations are first paired with their
modeled endpoint compliance ellipses. A Kinova Gen3 then prescribes a small circular displacement
$\Delta \mathbf{x}_{\mathrm{cmd}}$ through the ball-socket interface while the force-torque sensor
measures the resulting reaction force $\mathbf{f}_{\mathrm{meas}}$. Because the displacement has
constant radius, the measured force trace visualizes the local stiffness: the largest force occurs in
the least compliant direction, making the stiffness ellipse visually perpendicular to the compliance
ellipse. The probe is displacement-controlled, so it returns a reaction force for each prescribed displacement and therefore captures stiffness rather than compliance directly. To obtain the arm's empirical task-space compliance, a global optimizer fits the compliance matrix that best maps the measured forces to the prescribed displacements, so this compliance is computed from the data rather than measured directly. The analytical model maps the measured force back to displacement,
$\Delta \mathbf{x}_{\mathrm{model}}=C_{\mathbf{x}}\mathbf{f}_{\mathrm{meas}}$, and this modeled
displacement is compared with the commanded circular trajectory. Agreement between the modeled and
prescribed displacement ellipses verifies that the modeled Cartesian compliance captures the compliance
response of the physical arm.

\subsection{Simulated Compliance Validation}

The simulated compliance test uses the complementary protocol.
This protocol is included for two reasons. First, it validates the compliance map in the direction the
controller actually uses it, because the map sends an applied force to a tip displacement and so complements
the displacement-controlled hardware test that probes the inverse direction. Second, it confirms that the
simulator itself reproduces the modeled compliance, which is a prerequisite for trusting the later
simulation-based morphology study.
In simulation, the same McKibben actuator
model and robot geometry are tested by prescribing small external forces $\mathbf{f}_{\mathrm{cmd}}$
directly at the tip. For each of five test cases, a random pressure target is prescribed to the
actuators, and eight force directions are sampled around the local compliance ellipse. After the arm
settles, the resulting displacement
$\Delta \mathbf{x}_{\mathrm{sim}}$ is measured and compared with the analytical estimate
$\Delta \mathbf{x}_{\mathrm{model}}=C_{\mathbf{x}}\mathbf{f}_{\mathrm{cmd}}$. This force-controlled
test is practical in simulation because the applied tip force can be specified directly and repeatably,
which is difficult to achieve on the physical system.

\begin{figure}[t]
    \centering
    \includegraphics[width=\linewidth]{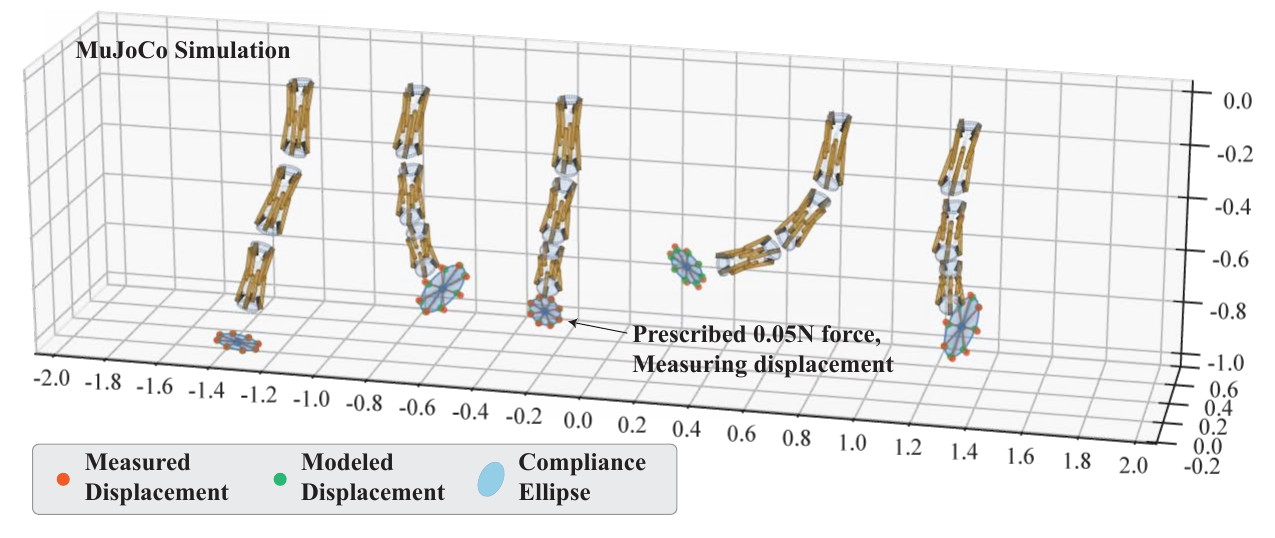}
    \caption{Simulated compliance validation against the analytical model. Five simulated
        force-displacement test cases are shown. In each case, eight prescribed tip forces sample the
        analytical model's local compliance ellipse; the simulated system provides the actual settled displacement, and the analytical
        model estimates the displacement by
        $\Delta \mathbf{x}_{\mathrm{model}}=C_{\mathbf{x}}\mathbf{f}_{\mathrm{cmd}}$. Agreement
        between the simulated and model-predicted displacement endpoints indicates that the modeled
        compliance captures the local force-to-displacement response.
        }
    \label{fig:mujoco_compliance_vs_modeling}
\end{figure}

Fig.~\ref{fig:mujoco_compliance_vs_modeling} summarizes the five simulated test cases. Sampling eight
force directions tests both the principal and intermediate directions of the local compliance profile.
Across the five cases, the simulated displacement endpoints remain close to the model-predicted
endpoints, showing that the analytical compliance model accurately predicts the simulated
force-to-displacement response in the tested neighborhoods.

Together, the real and simulated compliance tests evaluate the same local compliance map from opposite
directions: the physical system checks whether measured reaction forces map back to the imposed
displacement, while the simulated system checks whether imposed forces produce the predicted
displacement. Agreement in both cases verifies that the model accurately captures
the directional compliance needed by the controller.

\subsection{Simulation-to-Hardware Pressure-Target Posture Validation}

The compliance tests in Figs.~\ref{fig:real_system_compliance_probing} and
\ref{fig:mujoco_compliance_vs_modeling} verify the local compliance model, but agreement between
modeled and tested compliance alone is not sufficient to justify using the simulated system as a proxy for the physical
robot. The simulator must also respond similarly to the real pneumatic system when both are driven by
the same actuator-level command. We therefore record an actuator input sequence on the real arm and
replay the identical sequence in simulation, then compare the resulting joint-angle trajectories of the two
systems over the same time window.

\begin{figure}[t]
    \centering
    \includegraphics[width=\linewidth]{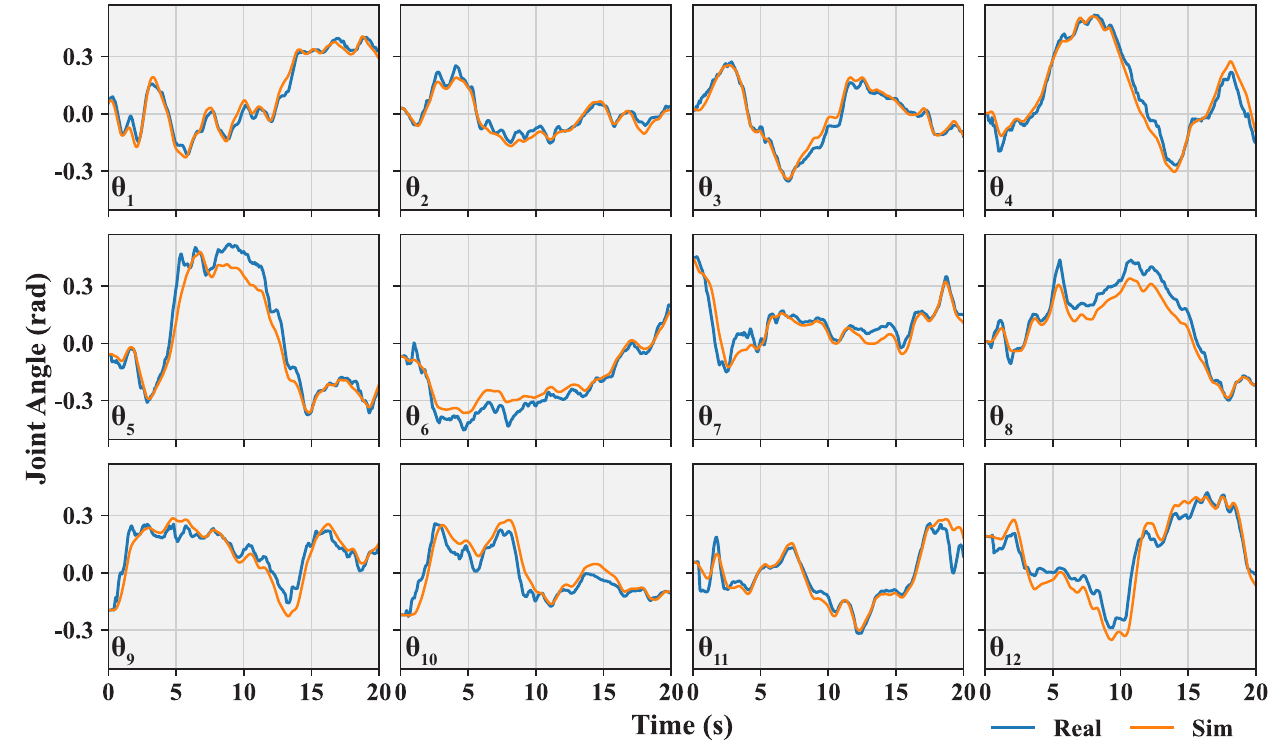}
    \caption{Simulation-to-hardware validation under a matched actuator input sequence. The input
        sequence recorded on the real arm is replayed in simulation, and the resulting joint-angle
        trajectories are compared head-to-head. Each of the twelve subplots corresponds to one entry of
        the joint-angle vector $\mathbf{q}\in\mathbb{R}^{12}$ (two angles per universal joint across the
        six joints) over a $20$~s window; the simulated trajectory (orange) is overlaid on the measured
        real-system trajectory (blue). The horizontal axis is time in seconds and the vertical axis is
        joint angle in radians. Across all twelve coordinates the simulation captures the dominant trend
        of the physical response, supporting the use of the simulated system as a behavioral proxy for the tested
        inputs.}
    \label{fig:pressure_target_joint_angle_comparison}
\end{figure}

Fig.~\ref{fig:pressure_target_joint_angle_comparison} overlays the simulated and physical joint-angle
responses under this matched input over a $20$~s window. The arm configuration is described by the
joint-angle vector $\mathbf{q}\in\mathbb{R}^{12}$, with two rotational degrees of freedom for each of
the six universal joints, so the figure contains one subplot per joint coordinate. In every subplot
the simulated trajectory (orange) is plotted directly against the measured real-system trajectory (blue),
with time in seconds on the horizontal axis and joint angle in radians on the vertical axis. Across all
twelve coordinates the simulated curves track the major trend of the physical curves with good
accuracy, capturing both the direction and the approximate magnitude of each joint's motion. This
confirms that the simulator reproduces the dominant pressure-to-motion behavior of the physical system
for the tested inputs. The remaining mismatch reflects effects such as friction, valve dynamics,
tubing compliance, and manufacturing variation.

\section[Model-Based Design for Isotropic Compliance]{Model-Based Design for Isotropic Compliance}
\label{sec:model_based_design}

Having established that the simulation reproduces the real arm's compliance and pressure-driven
posture, we now use it as a design tool, treating its predicted compliance as a faithful guide to the
real system and exploring a morphology that is not yet practical to manufacture.

A practical concern with the baseline arm is that its three segments are stacked coaxially, so at the
neutral pose $\mathbf{q}=\mathbf{0}$ the tip sits on a kinematic singularity. At this singularity the tip
cannot translate along the arm axis to first order, and one principal direction of the tip-compliance
ellipsoid collapses to zero compliance (infinite stiffness) for any actuation. We use the validated simulator to show that this
singular resting compliance is a consequence of how the segments are arranged, not of the modeling
framework or the actuators, and that it can be removed by \emph{reconfiguration}: rearranging the same
segments into a different fixed geometry while keeping the joint architecture, actuator model, and
pressure-control formulation unchanged.

Reconfiguration changes only segment lengths and the fixed orientation of each inter-segment connector.
A trial-and-error search over a wide range of candidate connector layouts selected a specific design that
equalizes the tip compliances along the three principal directions at $\mathbf{q}=\mathbf{0}$. In this
design the middle segment is lengthened, and the second and third connectors are pitched in opposite
directions ($+90^{\circ}$ and $-70^{\circ}$ about the world $x$-axis), so the arm folds into an S-curve
that is non-collinear at rest. Because the segments are no longer stacked, the resting-pose compliance singularity
disappears. As with the validation tests, compliance is evaluated through the same analytical model and
its simulated realization, with actuator pressures held at a gravity-free equilibrium so $\mathbf{q}=\mathbf{0}$
remains a static equilibrium.

\begin{figure}[t]
    \centering
    \includegraphics[width=\linewidth]{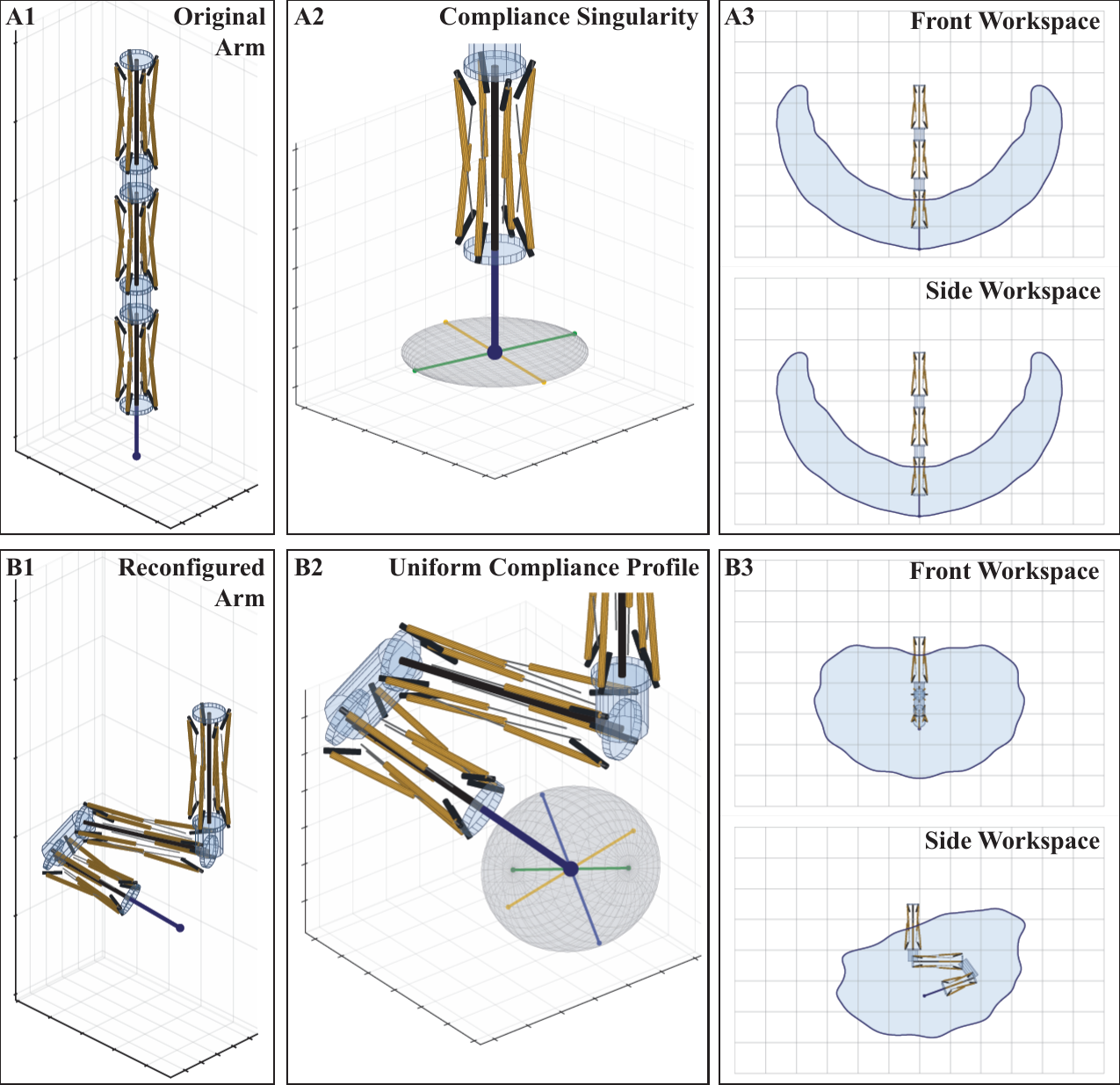}
    \caption{Reconfiguring the same segments removes the resting-pose compliance singularity. Row~A,
        baseline straight-stack arm; row~B, reconfigured pitch counter-return S-curve arm (same
        actuators; only segment lengths and connector orientations changed). Columns: (1) neutral-pose
        ($\mathbf{q}=\mathbf{0}$) skeleton; (2) tip-compliance ellipsoid at $\mathbf{q}=\mathbf{0}$ with
        pressures optimized for maximum isotropy (axis length $\propto$ compliance;
        blue/yellow/green~$=$~stiffest/middle/softest principal direction); (3) reachable-workspace
        cross-sections through the front ($x$--$z$) and side ($y$--$z$) symmetry planes (joints swept
        $\pm 40^{\circ}$). The baseline ellipsoid is a flat, singular lens (isotropy $\approx 0$; its
        zero-compliance axis is drawn with a small clamped thickness for visibility), while the
        reconfigured ellipsoid is near-isotropic (isotropy $0.88$).
        The reachable-workspace cross-sections show kinematic reachability under the joint-limit
        sweep and do not prune self-intersecting configurations, so some plotted points may correspond
        to poses that would self-collide on the physical arm; accounting for self-collision is left to
        future work.
        }
    \label{fig:reconfiguring}
\end{figure}

Fig.~\ref{fig:reconfiguring} compares the baseline (row~A) and reconfigured (row~B) arms on the
neutral-pose skeleton, the tip-compliance ellipsoid at $\mathbf{q}=\mathbf{0}$ optimized for maximum
isotropy, and reachable-workspace cross-sections in the front and side symmetry planes. The ellipsoids
show each arm's best achievable isotropy, defined here as the ratio of the smallest to the largest
principal tip compliance (so $1$ is a perfectly isotropic sphere and $0$ a degenerate, singular ellipsoid).
The baseline ellipsoid is a flat lens with one singular axis (isotropy $\approx 0$), whereas the
reconfigured ellipsoid is a near-uniform sphere with isotropy $0.88$; its principal stiffnesses are $12.8$, $11.9$, and $11.2$~N/m, equivalent compliances of $0.078$, $0.084$, and $0.089$~m/N. The reachable region changes correspondingly: the straight baseline
sweeps a hollow crescent below the base, while the reconfigured arm reaches a broad, filled lobe. These
panels are 2-D cross-sections through symmetry planes and compare the reachable shapes.

Two points connect this experiment to the paper's larger argument. First, the analytical compliance
model and its simulated realization apply unchanged to the reconfigured morphology, so the modeling and
analysis generalize across arm configurations rather than being tied to one build. Second,
reconfiguration is a design lever for task-appropriate compliance. The baseline's
near-singular compliance suits tasks that require large compliance switching in the $x$-$y$ plane, while
the reconfigured arm's omnidirectional compliance is better for reaching into unobservable spaces, where
the arm should yield safely in any direction to protect itself and its surroundings. Connected differently, the same segments
yield qualitatively different compliance and workspace. The validated model predicts
both before any new hardware is built.

\section{Hardware Demonstrations}
\label{sec:hardware_demonstrations}

The hardware demonstrations test the claim that quantitative structural compliance is useful when the
environment cannot be perfectly observed or held fixed. In both tasks, the arm must preserve useful tip
motion while allowing physical contact to correct disturbances that would be difficult to model as an
explicit trajectory.

\subsection{Dynamic Disturbance Rejection on a Moving Whiteboard}

The first demonstration is a writing task on a moving whiteboard. The robot must maintain the position
trajectory needed to write letters while the contact surface moves unpredictably relative to the arm.
In the moving-board trials, the Kinova Gen3 commands the whiteboard to move back and forth along a single
axis, which is the board-normal (pen-pressure) direction. The board moves at $1.5$~cm/s and $1.5$~Hz while
holding its orientation fixed, so the disturbance enters primarily along one axis.
With rigid behavior, the disturbance perturbs the contact force and degrades the drawn trajectory more
severely. With directional compliance enabled, the arm can absorb selected components of the
board motion through structural deflection while the IK controller continues to regulate the nominal
pen path. We deliberately do not use adaptive task-specific retuning in this experiment; the robot
uses the default position controller together with the selected inverse compliance profile so that the
effect of directional compliance can be isolated.

\begin{figure}[t]
    \centering
    \includegraphics[width=\linewidth]{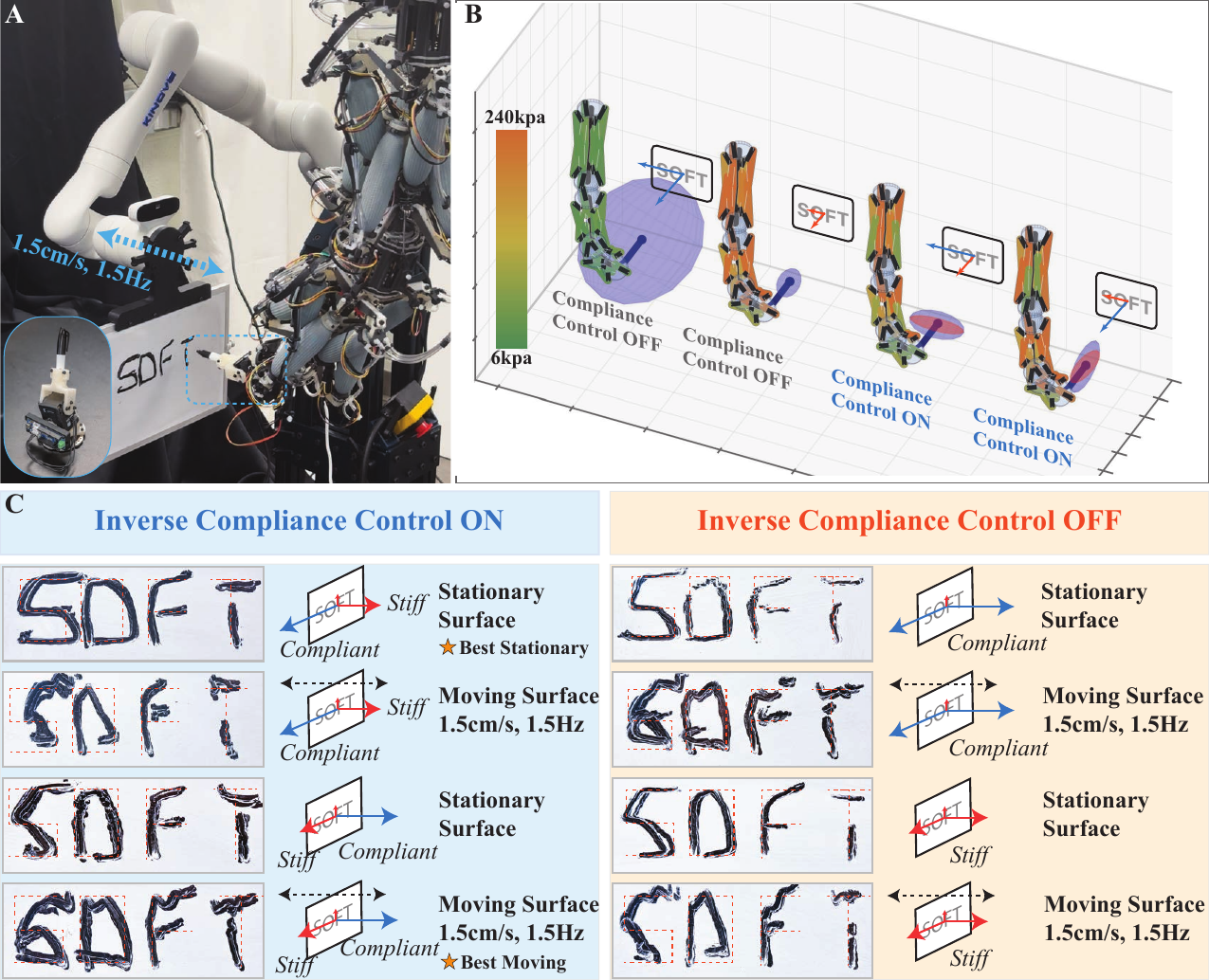}
    \caption{Dynamic writing demonstration on a moving whiteboard. (A) Experimental setup for
        writing the word ``soft.'' The whiteboard is held and moved by a Kinova Gen3 arm, while the
        rigid-soft arm writes through a 2-degree-of-freedom motorized pen end-effector module shown in the inset; the
        moving-board condition is marked as 1.5~cm/s at 1.5~Hz. (B) Skeleton-model reconstructions
        from motion-capture and pressure data for four compliance profiles: $x$-stiff/$y$-soft,
        $x$-soft/$y$-stiff, all stiff, and all soft. Actuator pressure is encoded by color. (C)
        Final written results for eight trials, with each compliance profile tested on stationary
        and moving surfaces. The left column shows inverse compliance control ON for the two
        directional profiles, and the right column shows compliance control OFF for the all-stiff and
        all-soft baselines.}
    \label{fig:UMArm_compliance_writing_with_disturbances}
\end{figure}

Fig.~\ref{fig:UMArm_compliance_writing_with_disturbances} shows the writing setup, the reconstructed
arm states, and the final written results for stationary and moving-surface trials.

The eight writing trials show that the useful compliance direction depends on how contact uncertainty
enters the task. For a stationary board, the $x$-stiff/$y$-soft profile produces the clearest letters:
high stiffness normal to the board maintains reliable pen pressure, while horizontal compliance reduces
jitter as the pen moves laterally. When the board moves, however, the same horizontal compliance becomes harmful, because the surface disturbance perturbs the pen sideways and produces inaccurate letters. The $x$-soft/$y$-stiff profile is less suitable for the stationary board because the pen pressure
is lower. It is more robust to board motion, however, because board-normal compliance passively absorbs the
disturbance while lateral stiffness preserves the letter shape. The compliance-control-off baselines
show the complementary failure modes: the all-soft arm struggles to maintain pen pressure, whereas the
all-stiff arm can write acceptably when the board is stationary but degrades strongly once the board
moves. The best stationary and moving-surface results both occur with inverse compliance control
enabled, demonstrating that actively selecting a directional compliance profile improves contact
writing through compliance alone.

\subsection{Unobservable Misalignment Correction in a Key-and-Drawer Task}

\begin{figure}[t]
    \centering
    \includegraphics[width=\linewidth]{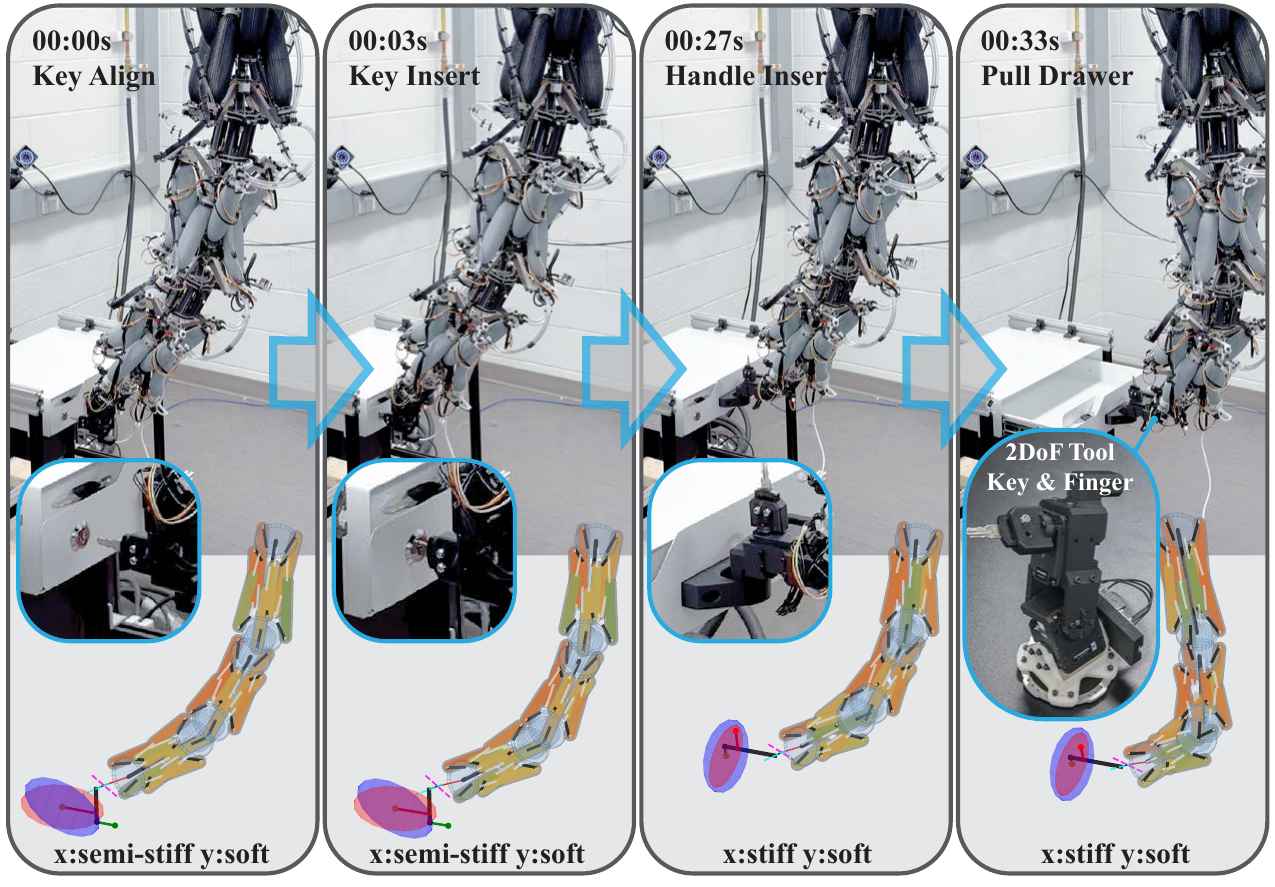}
    \caption{Real-time compliance tuning during a sequential key-unlock and drawer-opening task. A
        deliberately rigid end-effector is used so that the compliance observed during contact comes
        from the controlled arm structure rather than from a soft tool. The four columns are
        time-stamped snapshots of the task: (1) key alignment, (2) key insertion, (3) finger insertion
        into the handle, and (4) drawer pull-out. Each column shows, top to bottom, the real system, a
        picture-in-picture close-up at the key or handle, and the skeleton-model reconstruction with the
        target and achieved endpoint compliance ellipses plotted at the tip. During key insertion
        (columns~1--2) the profile is $x$:~semi-stiff, $y$:~soft: the soft axis along the insertion
        direction lets the key passively self-seat and correct the unobserved keyhole offset, while the
        semi-stiff orthogonal axis keeps the key on target. For drawer opening (columns~3--4) the
        profile switches to $x$:~stiff, $y$:~soft: the stiffened pulling axis transmits the opening force
        so the reaction load cannot push the rigid finger out of the handle, while the soft orthogonal
        axis absorbs residual misalignment as the drawer travels, protecting both the handle and the
        arm.}
    \label{fig:key_drawer_demonstration}
\end{figure}

The second demonstration is a sequential key-unlock and drawer-opening task that exhibits real-time
compliance tuning under unobservable contact geometry. The arm is fitted with a deliberately rigid
end-effector so that the compliance observed during contact originates from the controlled arm
structure rather than from a soft tool, isolating the contribution of the inverse compliance
controller. The robot is commanded to nominal key and handle locations, but the exact contact geometry
is not assumed to be perfectly known, so the controller relies on directional structural compliance to
passively resolve the residual misalignment. Fig.~\ref{fig:key_drawer_demonstration} shows four
time-stamped snapshots of the task: key alignment, key insertion, finger insertion into the handle, and
drawer pull-out.

During key insertion, the controller selects a soft profile along the insertion direction and a
semi-stiff profile in the orthogonal direction (Fig.~\ref{fig:key_drawer_demonstration}, columns~1--2,
$x$:~semi-stiff, $y$:~soft). The soft insertion axis lets the key yield and passively self-seat into the
keyhole, correcting the unobserved offset and easing entry, while the semi-stiff orthogonal axis keeps
the key from drifting off the hole.

After the drawer is unlocked, the end-effector switches to finger mode. The rigid finger first aligns
with the handle, and once it is engaged in the handle,
the controller stiffens the pulling direction while keeping the orthogonal direction soft
(Fig.~\ref{fig:key_drawer_demonstration}, columns~3--4, $x$:~stiff, $y$:~soft).
Opening the drawer requires the arm to follow a rigid external constraint, which is set by the drawer's
fixed line of travel. The arm must also exert enough force along the opening direction to overcome the
drawer's resistance. Stiffening the pulling axis lets the arm transmit this force, and it keeps the pulling
reaction from backing the rigid finger out of the handle. Meanwhile, the soft orthogonal axis lets the
end-effector yield sideways, so it can follow the exact path the drawer dictates.
A fully rigid arm has no such give. Any mismatch between its commanded path and the drawer's actual line
of travel produces large internal contact forces, which can jam the drawer, force the finger out of the
handle, or overload the arm. The tuned compliance instead absorbs that residual misalignment and protects
both the handle and the arm. This
sequence demonstrates why real-time compliance tuning matters: the robot moves from a soft
search-and-seat behavior to a stiffer load-bearing behavior with the same hardware, and successfully
completes the task without extensive path planning or sophisticated sensing.
\section{Discussion}
\label{sec:discussion}

The results support the central premise of this work. Quantitative structural compliance becomes easier to
control in real time when the hardware is deliberately designed to support iterative compliance and
kinematics control. The arm does not
remove the nonlinearities of pneumatic McKibben actuation, but it organizes them into a sparse model
that can be updated continuously. This makes the system suitable for tasks that require tracking a smooth
compliance and position trajectory.
The validation results confirm the sim-to-real agreement. The compliance tests show that the analytical
model predicts both the actual hardware displacement-to-force behavior and the simulated system
force-to-displacement behavior. The matched pressure-target test then shows that the simulated
system and the real arm produce similar joint-angle responses under the same actuator command. The simulation study then
extends to the hardware design: once the model and tested pressure-response behavior are
validated, segment geometry can be screened for its
achievable compliance before a difficult physical build.

\subsection{Known Hardware Limitations}

The current hardware still has important limitations. First, the baseline morphology exhibits a resting-pose compliance singularity at $\mathbf{q}=\mathbf{0}$, where one principal compliance along the arm axis collapses to near zero (very high stiffness), which limits how independently the
controller can tune vertical compliance. The reconfiguration study suggests that this limitation is not
inherent to the compliance-control formulation. Rearranging the same segments into a non-collinear
S-curve removes the resting-pose singularity in simulation, and it yields a near-isotropic tip compliance
(isotropy $0.88$, Fig.~\ref{fig:reconfiguring}).

When building the system, we were limited by the available off-the-shelf parts and our manufacturing
capabilities, so the mechanical parameters were not globally co-optimized. Link lengths, actuator
sizing, moment arms, and U-joint spacing all affect both workspace and compliance. We have shown that
better morphologies may exist, but our present capabilities prevent us from testing many of them. The
current arm should therefore be treated as a proof-of-concept hardware realization rather than an optimal
morphology.

\subsection{Known Algorithmic Limitations}

The compliance-Jacobian controller is a local method. It depends on the current configuration and
pressure state, and it follows the compliance gradient available at that state. As the benchmark shows,
this makes the method susceptible to local minima, so its mean static compliance error is higher than a
global optimizer's. When time is available, a global optimizer can therefore find lower-error static
solutions. In practice, however, the arm's redundancy keeps the local solution close enough for the task. The
trade-off is clear. Global optimization pays for that broader search with computation time and possible
discontinuity. The iterative method instead prioritizes immediate, smooth updates and lower energy use
during execution. A globally optimal variant of the iterative approach does exist. Rice et al. pair the
compliance-Jacobian solver with homotopy theory to reach a globally optimal compliance control strategy
\cite{rice2020homotopy}. We do not pursue it here,
but it points to a way to remove the local-minimum
limitation in future work.

\subsection{Proposed Solutions and Future Work}

Future controllers should use posture heuristics and secondary null-space tasks to avoid configurations
with compliance and kinematic singularities. The current null-space posture-relief idea can be
extended into an explicit objective that steers the arm away from singular compliance configurations
while preserving the end-effector task.

Future hardware should turn the simulated reconfiguration into a manufacturable design. The next design iteration can jointly optimize link lengths,
segment connection offsets, actuator routings, and joint spacing. The optimization target should be both
the reachable workspace and the achievable compliance.

Furthermore, the current compliance and position control rests on a quasistatic assumption. Developing a
compliance-dynamics controller is a promising direction.

Finally, global optimization and iterative methods should not be viewed as mutually exclusive. Global
optimization remains valuable for offline design, calibration, and static planning. The practical path
is to use global methods where time is available, and to use the compliance-Jacobian controller for the fast
contact response required by dynamic manipulation.
\section{Conclusion}
\label{sec:conclusion}

This paper presented a rigid-soft pneumatic arm and controller for real-time simultaneous position and
task-space compliance control. The hardware was designed around the needs of the algorithm. Six
mechanically isolated U-joints and independently regulated McKibben actuators create a sparse,
pressure-dependent stiffness model, which in turn supports local Jacobian-based control. The resulting controller
combines inverse kinematics, inverse compliance, null-space posture relief, and a dual-loop PID
controller for joint angle and joint energy.

Algorithmic benchmarks and hardware demonstrations show that this structure enables real-time
compliance changes during motion and contact. The compliance model is supported by real-system
displacement-to-force validation and simulated force-to-displacement validation. In addition, matched
pressure-target tests show that the simulator reproduces the dominant joint-angle response of
the physical system for the tested commands. Simulations then show how segment length and
connection offset can be tuned to enlarge the attainable compliance workspace in future hardware. Using
the physical system we built,
we demonstrated that tasks normally requiring sophisticated sensing and control can instead be accomplished through structural compliance.
More broadly, the system demonstrates the benefits of utilizing hyper-redundancy and incorporating a rigid skeletal structure into a soft robotics framework.
Used deliberately, the redundancy and the rigid skeleton make soft actuation easier to model and faster to control. They also
make soft actuation more useful in unstructured environments where precision and compliance are both
required.

\bibliographystyle{IEEEtran}
\bibliography{references}

\end{document}